\pdfoutput=1  
\PassOptionsToPackage{table}{xcolor}
\documentclass[10pt]{article}

\usepackage[top=2.25cm,bottom=2.5cm,left=2.5cm,right=2.5cm,columnsep=0.65cm]{geometry}
\usepackage{microtype}
\usepackage{fancyhdr}
\usepackage{placeins}
\usepackage{parskip}
\usepackage[english]{babel}
\usepackage{etoolbox}
\usepackage{graphicx}
\usepackage{float}
\usepackage{subcaption}
\usepackage{booktabs}
\usepackage{multirow}
\usepackage{amsmath}
\usepackage{amssymb}
\usepackage[most]{tcolorbox}
\usepackage[T1]{fontenc}
\usepackage[utf8]{inputenc}
\usepackage{lmodern}
\usepackage{url}
\usepackage{enumitem}
\usepackage{needspace}
\usepackage{tabularx}
\usepackage{colortbl}
\usepackage{tikz}
\usetikzlibrary{arrows.meta,positioning}

\usepackage{titlesec}
\titleformat*{\section}{\large\bfseries}
\titleformat*{\subsection}{\normalsize\bfseries}
\titleformat*{\subsubsection}{\normalsize\bfseries}
\titleformat*{\paragraph}{\normalsize\bfseries}

\usepackage{caption}
\usepackage[numbers,sort&compress]{natbib}

\definecolor{darkorange}{HTML}{d4792a}
\definecolor{cInk}{HTML}{4A4A4A}
\definecolor{cBlue}{HTML}{3A6EA5}
\definecolor{cBlueBg}{HTML}{E4EEF7}
\definecolor{cGreen}{HTML}{2E7D5B}
\definecolor{cGreenBg}{HTML}{E0F0E7}
\definecolor{cRed}{HTML}{A63A3A}
\definecolor{cRedBg}{HTML}{F7E4E4}

\usepackage[colorlinks,linkcolor=black,citecolor=darkorange,urlcolor=darkorange]{hyperref}
\hypersetup{pdftitle={When Can a Machine Trust a Statute? A Survival Certificate for Machine-Extracted Legal Logic},pdfauthor={Surya Saka},pdfsubject={Noise-calibrated certification of statutory implication bases}}

\newcommand{\papershorttitle}{When Can a Machine Trust a Statute?}
\fancypagestyle{main}{%
  \fancyhf{}%
  \fancyhead[C]{\small\itshape\papershorttitle}%
  \fancyfoot[R]{\thepage}%
  }
\fancypagestyle{firststyle}{%
  \fancyhf{}%
  }
\newcolumntype{L}[1]{>{\raggedright\arraybackslash}p{#1}}
\newcommand{\dstar}{D$^{\star}$}
\newcommand{\attr}[1]{\textsc{#1}}
\newcommand{\exact}{\textsc{exact}}
\newcommand{\tolerant}{\textsc{tolerant}}

\newtcolorbox{keybox}[1]{colback=cBlueBg,colframe=cBlue,arc=1pt,
  left=8pt,right=8pt,top=6pt,bottom=6pt,title={#1}}
\newtcolorbox{warnbox}[1]{colback=cRedBg,colframe=cRed,arc=1pt,
  left=8pt,right=8pt,top=6pt,bottom=6pt,title={#1}}
\newtcolorbox{okbox}[1]{colback=cGreenBg,colframe=cGreen,arc=1pt,
  left=8pt,right=8pt,top=6pt,bottom=6pt,title={#1}}

\title{\bfseries When Can a Machine Trust a Statute?\\[2pt]
A Survival Certificate for Machine-Extracted Legal Logic}

\author{Surya~Saka\\
  JudicialMind\\
  \url{surya@judicialmind.ai}}

\date{}

\begin{document}

\maketitle
\thispagestyle{firststyle}

\begin{abstract}
Statutes are increasingly parsed by machines before people read them, and the
parsers disagree: on Missouri's statutes, two independently written extractors diverge on
numeric-threshold presence at a false-negative rate of 0.43. We ask what
formal logic survives such noise. We build a passive survival certificate for
the Duquenne--Guigues implication basis of machine-extracted statutory
contexts: per-attribute inter-extractor disagreement is measured, replayed
against the basis in 1{,}000 Monte Carlo trials, and an implication is
certified only when a one-sided Wilson 95\% lower bound on survival reaches
0.95; every certified implication carries premise spans and a minimal
counterexample. On 29{,}365 Missouri sections and 502 Indian central-Act
sections, the preregistered held-out gate passes (10 statute families across
7 Titles exact; 16 across 11 with 5\% tolerance), yet under one globally
deployed error model 93.2\% of held-out chapters fall below the
informativeness floor, and a $2\times2$ factorial assigns that to
calibration-rate transfer, not selection. The certificate is usable but
fragile: deploy it per-chapter-calibrated or error-tolerant. Code, data
products, and the audit trail, including one retracted claim, are released.

\end{abstract}

\section{Introduction}
\label{sec:intro}

Machine extraction is the first step of every statutory pipeline we know: a
parser reads the text and writes down structured attributes, and downstream
symbolic logic treats those attributes as exact. They are not exact. On the
pooled Missouri context used in this study, two rule-based parsers written
independently of each other disagree about the presence of a numeric
threshold at a false-negative rate of 0.43. The disagreement is not noise at
the margins; it sits on attributes that legal logic depends on.

The problem this creates is concrete, and one pair of implications from our
own results states it better than any framing. Over 2,555 pooled Missouri
sections, the canonical implication basis contains
$\{$\attr{deadline}$\} \to \{$\attr{numthresh}$\}$ with support 792: nearly
eight hundred sections state a deadline as a numeric duration, and the
implication looks like the safest rule in the corpus. Under the measured
disagreement model its survival probability is 0.000, because the second
extractor misses spelled-out numeric thresholds almost half the time and each
miss manufactures a counterexample. The same basis contains
$\{$\attr{oblig}, \attr{proviso}, \attr{unless}, \attr{numthresh},
\attr{penalty}, \attr{deadline}$\} \to \{$\attr{prohib}$\}$ with support 12,
an unglamorous regularity about heavily conditioned regulatory sections, and
its survival probability is 0.965, high enough to certify, because the
attributes it depends on are ones the two extractors almost never delete.
Support, the quantity every symbolic pipeline can see, does not predict
robustness, the quantity that matters at deployment. What predicts robustness
is the interaction between an implication's attribute footprint and the
measured error structure of the extractor.

This paper builds the instrument that makes that interaction explicit: a
\emph{survival certificate} for the Duquenne--Guigues (DG) implication basis
of a machine-extracted statutory context. Two extractors, a reference and an
independently authored token-adjacency parser, are run over the same statutes.
Their per-attribute disagreement calibrates a conditional error model. That
model is replayed against the reference basis in $N{=}1000$ Monte Carlo
trials, and each canonical implication receives a survival probability under
measured noise. An implication is certified when a one-sided Wilson 95\%
lower bound on survival reaches $\tau{=}0.95$ \citep{wilson1927probable}, with
a Benjamini--Hochberg companion for the certified set
\citep{benjamini1995controlling}. Every certified implication carries
provenance: premise and consequent spans, a minimal counterexample, and the
calibration-linked most likely way it breaks.

Three empirical findings organize the paper. First, support is not a reliable
proxy for robustness of machine-extracted legal implications; the certificate
separates implications that look identical under any support- or
confidence-based view (\S\ref{sec:res-worked}). Second, global calibration is
fragile because extraction disagreement is heterogeneous across statutory
chapters: under one pooled error model shipped to every chapter, 93.2\% of
held-out Missouri chapters fall below the informativeness floor, and a
preregistered $2{\times}2$ factorial assigns the failure to calibration-rate
transfer across heterogeneous chapters, not to held-out selection
(\S\ref{sec:res-factorial}). Third, per-chapter calibration materially
improves informativeness, lifting the held-out mean certified fraction from
0.038 to 0.128, while the instrument remains conditional and fragile: it is a
statistical lower bound on survival under an empirically measured
inter-extractor disagreement model, not a guarantee of legal correctness
(\S\ref{sec:res-local}, \S\ref{sec:discussion}).

The controlling evaluation is preregistered and held out. The gate requires
the certificate to be informative on at least two held-out Missouri families
from at least two distinct Titles, with the error model measured on a
disjoint calibration half. It passes: 10 eligible chapters across 7 Titles
under exact semantics, 16 across 11 under a 5\% error-tolerant semantics. The
cross-jurisdiction replication fails: on IndiaCode, 0 of 6 eligible Acts land
in the informative band. We report both without smoothing either.

Contributions. (i) A passive survival certificate for the DG basis over
machine-extracted statutory contexts: per-implication survival under a
measured conditional error model, Wilson lower-bound certification at a
common Monte Carlo budget, exact and $\varepsilon$-tolerant semantics, and a
premise-collapse/violation decomposition (\S\ref{sec:method}). (ii) A
counterexample-carrying provenance layer that makes every certified
implication auditable by hand (\S\ref{sec:provenance}). (iii) A
preregistered held-out evaluation on Missouri and IndiaCode, with a factorial
deconfounding of selection from calibration-rate transfer, negative controls,
and a 458-chapter scale run (\S\ref{sec:setup}, \S\ref{sec:results}).

Scope. The certificate bounds the probability that an implication survives
parser disagreement. It says nothing about whether the parser is legally
correct, whether the reference extraction is ground truth, or whether a
certified implication is a real legal rule. The distinction between
$\Pr[\text{implication survives measured disagreement}]$ and
$\Pr[\text{extraction is legally correct}]$ is maintained throughout, and
\S\ref{sec:limitations} states the limits of the instrument exactly.

\section{Related Work}
\label{sec:related}

\paragraph{FCA and implication bases.}
Formal Concept Analysis turns a binary object--attribute table into a concept
lattice; the Duquenne--Guigues basis is its canonical, minimal, complete
implication set \citep{ganter1999fca}. The basis's combinatorics are settled
\citep{kuznetsov2008decision} and enumeration is fast in practice
\citep{janostik2021lincbo}. This literature assumes a correct context. We
study what happens to the basis when the context is machine-extracted and
measurably noisy.

\paragraph{Noisy and approximate implication bases.}
Approximate bases of exact association rules \citep{kanda2001constructing},
approximate computation of exact rules \citep{bansal2021approximate}, and
confidence-based redundancy removal \citep{balcazar2013confidence} relax the
object to gain robustness; the robustness of full-implication inference has
also been analysed directly \citep{dai2013robustness}. Our
$\varepsilon$-tolerant semantics follows Borchmann's error-tolerant
construction of $\mathcal{EL}^{\bot}$ ontologies \citep{borchmann2013towards}.
These works deliver a new, approximate object. We keep the exact canonical
basis as the artifact and ask which of its members survive a measured
perturbation, with $\varepsilon$ treated as a declared sensitivity parameter.

\paragraph{Stability and PAC-style robustness.}
Concept stability ranks concepts by resistance to object deletion
\citep{babin2012approximating,buzmakov2014scalable}; it uses no noise model
and answers a resampling question, not an extraction question. PAC
implication bases carry distributional learning guarantees
\citep{borchmann2017usability}. Our guarantee is per-implication and
frequentist: a Wilson lower bound on survival under the calibrated error
process, with set-level multiplicity handled by Benjamini--Hochberg
\citep{wilson1927probable,benjamini1995controlling}.

\paragraph{Interactive error correction.}
The closest line corrects implications over erroneous contexts through expert
interaction \citep{kuznetsov2015interactive} and uses canonical implications
to flag errors in corrupted rows \citep{revenko2012finding}. Both assume a
human adjudicator. We replace adjudication with a measured error model and
certify passively; our counterexample carrier points the opposite way from
\citep{revenko2012finding}, explaining how a certified implication would
break rather than convicting a corrupt row.

\paragraph{Legal and statutory NLP.}
Legal FCA applications are sparse; the closest we could verify applies
concept lattices to Brazil's LGPD without modelling extraction noise
\citep{martins2020lgpd}. Production legal tools operate above the formal
layer and compute no implication bases, so there is nothing in that ecosystem
to certify against.

The gap this paper occupies: robustness of exact symbolic implications
induced from machine-extracted legal attributes, under measured extraction
disagreement, with an auditable counterexample and provenance layer.

\section{Method}
\label{sec:method}

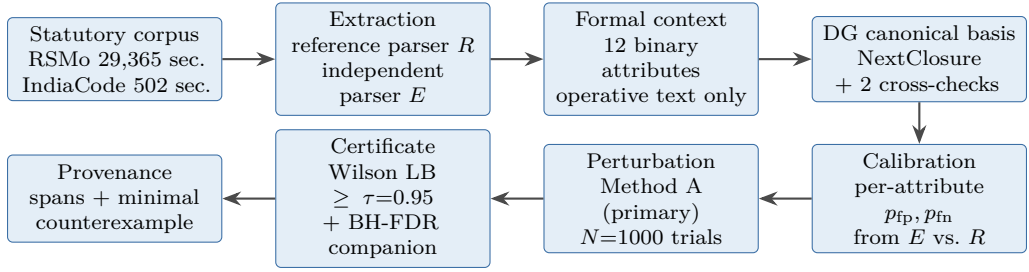
\begin{figure}[t]
\centering
\begin{tikzpicture}[
  node distance=0.55cm and 0.7cm,
  stage/.style={draw=cBlue, fill=cBlueBg, rounded corners=2pt,
                minimum height=1.0cm, text width=2.6cm, align=center,
                font=\footnotesize},
  arr/.style={-{Stealth[length=2.5mm]}, thick, color=cInk}]
\node[stage] (corpus) {Statutory corpus\\ RSMo 29{,}365 sec.\\ IndiaCode 502 sec.};
\node[stage, right=of corpus] (extract) {Extraction\\ reference parser $R$\\ independent parser $E$};
\node[stage, right=of extract] (context) {Formal context\\ 12 binary attributes\\ operative text only};
\node[stage, right=of context] (basis) {DG canonical basis\\ NextClosure\\ + 2 cross-checks};
\node[stage, below=of basis] (calib) {Calibration\\ per-attribute $p_{\mathrm{fp}}, p_{\mathrm{fn}}$\\ from $E$ vs.\ $R$};
\node[stage, left=of calib] (perturb) {Perturbation\\ Method A (primary)\\ $N{=}1000$ trials};
\node[stage, left=of perturb] (cert) {Certificate\\ Wilson LB $\geq \tau{=}0.95$\\ + BH-FDR companion};
\node[stage, left=of cert] (prov) {Provenance\\ spans + minimal\\ counterexample};
\draw[arr] (corpus) -- (extract);
\draw[arr] (extract) -- (context);
\draw[arr] (context) -- (basis);
\draw[arr] (basis) -- (calib);
\draw[arr] (calib) -- (perturb);
\draw[arr] (perturb) -- (cert);
\draw[arr] (cert) -- (prov);
\end{tikzpicture}
\caption{\textbf{Pipeline.} The reference context is documentary truth by
construction; calibration measures how a second, independently authored parser
disagrees with it; certification asks which canonical implications survive
that measured disagreement.}
\label{fig:pipeline}
\end{figure}

Figure~\ref{fig:pipeline} shows the pipeline end to end. This section defines
each stage; implementation detail that does not affect the scientific
argument is deferred to the appendices.

\subsection{Binary statutory context}
\label{sec:context}

A formal context $\mathbb{K} = (G, M, I)$ has sections $G$ as objects, binary
attributes $M$, and incidence $I \subseteq G \times M$
\citep{ganter1999fca}. For $X \subseteq M$ the derivation
$X' = \{g \in G : g\mathrel{I}m \ \forall m \in X\}$ and its dual give the
closure operator $X''$; an implication $X \to Y$ holds in $\mathbb{K}$ iff
$Y \subseteq X''$, i.e.\ every section carrying all of $X$ also carries all
of $Y$. The support of $X$ is $\mathrm{supp}(X) = |X'|$.

The schema is frozen at 12 machine-verifiable attributes over operative text:
\attr{oblig}, \attr{prohib}, \attr{permis}, \attr{proviso}, \attr{unless},
\attr{except}, \attr{xref}, \attr{xref\_res}, \attr{def}, \attr{numthresh},
\attr{penalty}, \attr{deadline} (definitions in
Appendix~\ref{app:schema}). The reference context produced by the reference
parser is treated as documentary truth inside the experiment. Everything we
measure is robustness of downstream logic to extraction disagreement, not
correctness of the extraction itself.

\subsection{Duquenne--Guigues basis}
\label{sec:basis}

We compute the DG canonical basis with NextClosure \citep{ganter1999fca} and
cross-check it against an independent definition-based pseudo-intent oracle
and the \texttt{concepts} library; the three agree on every real context.
Enumeration is sub-second and bounded (max basis 59 over the nine headline
contexts, 62 over the 304/458-chapter sweeps, against a preregistered cap of
10{,}000). The evaluation target is the set of implications with premise
support at least $S_{\min}{=}1$; zero-support implications are dropped, and
the schema tautology $\attr{xref\_res} \to \attr{xref}$ is tagged and
reported separately.

\subsection{Calibration: measured inter-extractor disagreement}
\label{sec:calibration}

The error model is measured, not assumed. A second parser, written
independently of the reference (token-adjacency scanning, separately authored
vocabulary, negation-window polarity, digit and spelled-number recognition,
no shared pattern table), is run over the same sections. For each attribute
$m$ we record the conditional disagreement rates
$p_{\mathrm{fp}}(m) = \Pr[m \in E(g) \mid m \notin R(g)]$ and
$p_{\mathrm{fn}}(m) = \Pr[m \notin E(g) \mid m \in R(g)]$.
Table~\ref{tab:confusion} (Appendix~\ref{app:confusion}) gives the pooled
matrix: \attr{numthresh} dominates false negatives
($p_{\mathrm{fn}}{=}0.429$), while \attr{prohib}, \attr{xref},
\attr{except}, \attr{deadline} have measured $p_{\mathrm{fn}} = 0$.

This quantity is \emph{inter-extractor disagreement between two hand-built
parsers}. It is not externally validated extraction error, and the two
parsers share the upstream operative-text split and schema, so disagreement
plausibly understates real error (\S\ref{sec:limitations}). The certificate
is therefore conditional: it bounds survival under the measured disagreement
model, nothing more.

\subsection{Perturbation model}
\label{sec:estimators}

\textbf{Method A (primary).} Independent per-attribute conditional flips: in
each Monte Carlo trial, cell $(g, m)$ is added with probability
$p_{\mathrm{fp}}(m)$ if absent and removed with probability
$p_{\mathrm{fn}}(m)$ if present. Method A generates genuinely new errors and
assumes per-attribute independence; that assumption is disclosed and
stress-tested, not hidden.

\textbf{Method \dstar{} (joint-error diagnostic).} A polarity-conditional
joint residual bootstrap: whole observed per-section error vectors are drawn
from the calibration pool and applied cell-wise only where donor and target
reference polarity match, preserving joint co-error structure under the
measured conditional rates. \dstar{} exists to test whether Method A's
independence assumption understates damage; it is not a second independent
extractor. Mechanics and the invalid estimator it replaces are documented in
Appendix~\ref{app:audit}.

\subsection{Survival and certification}
\label{sec:certificate}

Given a perturbed context $\tilde{\mathbb{K}}$, implication $X \to Y$
\emph{survives non-vacuously} iff some row still supports $X$ and the
implication holds on all supported rows (\exact{}, $\varepsilon{=}0$) or on
at least $1{-}\varepsilon$ of them (\tolerant{}, primary
$\varepsilon{=}0.05$; \citealp{borchmann2013towards}). We additionally record
$S_{\mathrm{logic}}$, where vacuous retention counts as survival, and
decompose non-survival into \emph{premise collapse} (noise erases every
supporting row) versus \emph{instantiated violation} (a supported
counterexample appears). The distinction matters operationally: a violation
is checkable, because the killing row exists and the certificate names it; a
collapse is a support artifact.

Over $N{=}1000$ trials the survival estimate is
$\hat{p} = \#\text{survivals}/N$, and an implication is certified iff the
one-sided Wilson 95\% lower bound on $\hat{p}$ reaches $\tau{=}0.95$
\citep{wilson1927probable}:
\begin{equation}
\mathrm{LB}(\hat{p}, N) =
\frac{\hat{p} + \frac{z^2}{2N} - z\sqrt{\frac{\hat{p}(1-\hat{p})}{N} +
\frac{z^2}{4N^2}}}{1 + \frac{z^2}{N}} \;\geq\; \tau,
\label{eq:wilson}
\end{equation}
with $z{=}1.645$. The decision rule is budget-dependent (the minimum
observable survival that certifies falls from 0.980 at $N{=}200$ to 0.962 at
$N{=}1000$), so $N$ is part of the disclosed rule, fixed at 1000 for every
reported comparison and varied only in the sensitivity sweep
(Appendix~\ref{app:sensitivity}). For claims about certified \emph{sets} we
add a Benjamini--Hochberg FDR ($q{=}0.05$) certified fraction
\citep{benjamini1995controlling}. The second preregistered outcome is the
whole-basis semantic edit distance between reference and trial bases,
averaged over trials; it measures basis-level stability rather than
implication-level survival, and \S\ref{sec:discussion} keeps the two
quantities separate.

\subsection{Counterexample and provenance carrier}
\label{sec:provenance}

Every evaluated implication carries: premise and consequent text spans; a
minimal counterexample, the cheapest perturbation producing a supported
violating row; a calibration-linked likely counterexample, the counterexample
whose flipped attributes carry the highest measured disagreement; observed
counterexamples from the independent extractor's own context; and colocated
exception markers, recorded as proximity signals only. The purpose is audit
cost: a reviewer can falsify a certified implication against the statute text
against the statute text, and a fragile implication fails with a named witness.

\section{Experimental Setup}
\label{sec:setup}

\paragraph{Corpora.}
The primary corpus is the Missouri Revised Statutes: 29{,}365 section entries
(29{,}350 unique section ids; 15 duplicate files in chapters 1--3, disclosed)
across 458 chapters. Title membership is recovered from the official
Title$\to$chapter structure, verified to reproduce all eight design-family
pairs before use. The replication corpus is 15 Indian central Acts from
IndiaCode, 502 parsed sections, with act-scoped cross-reference resolution.

\paragraph{Design and held-out split.}
The exploratory headline pools 2{,}555 sections from 8 highlighted chapters
spanning 8 Titles; these families were selected during development, so
in-sample numbers are descriptive only. The \emph{controlling} evaluation is
held out: evaluation chapters exclude the 8 design families (148 evaluated,
146 eligible under the min-eval filter), the error model is measured on a
disjoint calibration half and transferred to every evaluation chapter, and
IndiaCode is a held-out jurisdiction. This design is controlling because it
is the only configuration that answers the deployment question: does the
certificate stay informative on statute families the method was not tuned on,
under an error model estimated elsewhere?

\paragraph{Scale views.}
Beyond the held-out design, an anti-cherry-pick sweep covers all 304 chapters
with at least 30 sections (26{,}787 sections), and a full-corpus scale run covers
all 458 chapters under the primary extractor, including a stability curve
over a noise-scaling grid.

\paragraph{Protocol constants.}
All confirmatory runs share the Monte Carlo budget $N{=}1000$ and pinned
seeds (headline seed 0; variance study $\{0,1,2\}$). Certification uses
$\tau{=}0.95$ with the Wilson lower bound (Eq.~\ref{eq:wilson}) and a
BH-FDR ($q{=}0.05$) companion. The informative band is $[0.10, 0.90]$:
below the floor the certificate says nothing useful, above the ceiling it
certifies nearly everything and stops discriminating. The preregistered
controlling gate: certified fraction in band on at least 2 held-out families
from at least 2 distinct Titles, each with $n_{\mathrm{eval}} \geq 10$, basis
bounded below 10{,}000. A stricter majority-in-band criterion from an earlier
revision is reported as a post-hoc prevalence diagnostic only
(Appendix~\ref{app:audit}).

\paragraph{Metrics.}
Per context we report: certified fraction (Wilson and FDR) under \exact{} and
\tolerant{} semantics, mean survival, premise-collapse share, and mean
whole-basis edit distance. Per implication: survival probability and the
provenance payload of \S\ref{sec:provenance}.

\section{Results}
\label{sec:results}

The results are organized around the scientific questions, not around every
experiment performed. The in-sample headline table, full $\varepsilon$ grid,
per-attribute confusion, support strata, per-Act India rows, ablations, and
seed variance are in the appendices and regenerate from the released
artifact.

\subsection{Support does not predict survival}
\label{sec:res-worked}

The motivating contrast of \S\ref{sec:intro} is a real pair of basis members
from the pooled Missouri context:

\begin{warnbox}{High support, zero survival}
$\{$\attr{deadline}$\} \to \{$\attr{numthresh}$\}$, support 792, survival
0.000. Deadlines are nearly always expressed numerically, so the rule looks
safe. But \attr{numthresh} carries the largest measured false-negative rate
in the schema (0.429): the second extractor misses spelled-out thresholds
nearly half the time, and every miss on one of 792 supporting rows is an
instantiated counterexample. The certificate's likely-counterexample pointer
names \attr{numthresh} with its measured rate and cites \S115.004
(``twenty-six weeks''); the minimal counterexample needs a single
\attr{deadline} insertion on \S115.003.
\end{warnbox}

\begin{okbox}{Low support, certified}
$\{$\attr{oblig}, \attr{proviso}, \attr{unless}, \attr{numthresh},
\attr{penalty}, \attr{deadline}$\} \to \{$\attr{prohib}$\}$, support 12,
survival 0.965, above the 0.962 certification cutoff at $N{=}1000$. Heavily
conditioned regulatory sections also prohibit. Its consequent sits on
\attr{prohib}, whose measured false-negative rate is exactly 0: the measured
noise almost never deletes a prohibition, so the implication is structurally
hard to kill. Its cheapest death costs one flip (add \attr{proviso} to
\S115.631, where \attr{prohib} is already absent).
\end{okbox}

A support- or confidence-based ranker cannot separate these two rules. The
survival certificate separates them, and the separation is explained, not
oracular: each outcome points at the attribute and the section responsible.

\subsection{Does the certificate survive on held-out Missouri?}
\label{sec:res-heldout}

Yes, at the preregistered bar. Under the controlling held-out design, the
\exact{} certificate is informative on \textbf{10 of 146} eligible chapters
spanning \textbf{7 distinct Titles} (IX, XI, XV, XVII, XXXVI, XL, XLI); the
\tolerant{} certificate on \textbf{16 of 146} across 11 Titles
(Table~\ref{tab:heldout}). The controlling gate (${\geq}2$ families,
${\geq}2$ Titles, $n_{\mathrm{eval}} \geq 10$, band $[0.10, 0.90]$) passes
on both semantics.

The same table carries the caution: the mean \exact{} certified fraction
across held-out chapters is 0.038 (chapter-bootstrap 95\% CI
[0.033, 0.044]), and 93.2\% of chapters fall below the floor (Wilson 95\%
[0.879, 0.962]). The gate passes and the naive deployment mode is weak. Both
are true, and the next subsection explains the weakness.

\begin{table}[t]
\centering\small
\begin{tabular}{@{}lcc@{}}
\toprule
 & Missouri & India \\
\midrule
contexts (eval / target / min-eval) & 148 / 148 / 146 & 8 / 8 / 6 \\
\exact{} mean certified fraction & 0.038 & 0.041 \\
\exact{} in band $[0.10,0.90]$ & \textbf{10/146} (7 Titles) & 0/6 \\
\exact{} below floor & 93.2\% & 6/6 \\
\tolerant{} ($\varepsilon{=}.05$) in band & \textbf{16/146} (11 Titles) & 0/6 \\
\dstar{} \exact{} in band & 44/146 & --- \\
\bottomrule
\end{tabular}
\caption{\textbf{Held-out confirmatory results} (common $N{=}1000$).
Evaluation chapters exclude the 8 design families; the error model is
measured on a disjoint calibration half and transferred; India is a held-out
jurisdiction. One excluded India Act carries a non-trivial 0.25 certificate
at $n_{\mathrm{eval}}{=}8$; denominators are shown before the min-eval
filter.}
\label{tab:heldout}
\end{table}

\begin{table}[t]
\centering\small
\begin{tabular}{@{}lccc@{}}
\toprule
configuration & pooled & held-out in band & verdict \\
\midrule
indep.\ + A + \exact{} (primary) & 0.067 & 10 / 7 Titles & \textbf{PASS} \\
indep.\ + A + \tolerant{}(.05)   & 0.633 & 16 / 11 Titles & \textbf{PASS} \\
indep.\ + \dstar{} + \exact{}    & 0.033 & 44 in band & consistent \\
\texttt{shallow} + A + \exact{}  & 0.200 & --- & disclosed artifact \\
\bottomrule
\end{tabular}
\caption{\textbf{Gate table} (common $N{=}1000$). ``Pooled'' is the
in-sample pooled context, descriptive only. The controlling criterion
(${\geq}2$ held-out families from ${\geq}2$ distinct Titles, each
$n_{\mathrm{eval}}{\geq}10$, band $[0.10,0.90]$) passes on \exact{} and
\tolerant{}. The \texttt{shallow} row is an engineered stress fixture that
shares code with the reference on 5/12 attributes; it is shown for contrast
and never gates.}
\label{tab:gate}
\end{table}

\subsection{Why does global deployment fail?}
\label{sec:res-factorial}

\begin{figure}[t]
\centering
\includegraphics[width=0.62\textwidth]{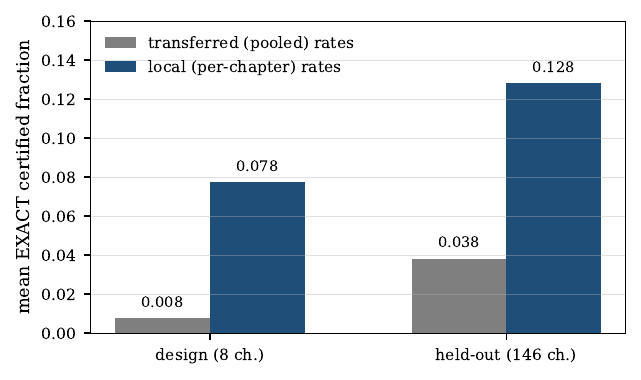}
\caption{\textbf{The $2{\times}2$ as a picture.} Switching the calibration
source lifts the mean certified fraction 3--10$\times$ on both family sets;
switching the family set at a fixed calibration source changes little, and in
the wrong direction for a selection story.}
\label{fig:factorial}
\end{figure}

The natural first explanation of the 93.2\% below-floor figure is
generalization failure: the method was tuned on the design families and
degrades off them. The preregistered $2{\times}2$ factorial
(Table~\ref{tab:factorial}, Figure~\ref{fig:factorial}) tests this by varying
family set \{design, held-out\} against calibration source \{transferred
pooled, local per-chapter\} at common $N$.

The verdict is clean. Moving from transferred to local calibration lifts the
mean certified fraction by $+0.090$ on held-out chapters and $+0.070$ on
design chapters: a large effect, similar in both rows. Moving from design to
held-out families at a fixed calibration source changes the mean by
$-0.031$ (transferred) and $-0.051$ (local): small, and negative, meaning
design chapters certify slightly \emph{lower} than held-out ones once the
calibration source is held fixed. The below-floor prevalence is therefore a
property of the global-deployment estimand, one pooled error model shipped
everywhere, and not a held-out selection effect.

The mechanism is chapter-level heterogeneity of the disagreement process.
Across chapters, \attr{numthresh} $p_{\mathrm{fn}}$ has mean $0.499 \pm
0.217$, \attr{proviso} $p_{\mathrm{fn}}$ $0.152 \pm 0.204$, \attr{oblig}
$p_{\mathrm{fp}}$ $0.138 \pm 0.153$. A single transferred rate set cannot
represent 146 chapters whose error structure varies at that amplitude.

\begin{table}[t]
\centering\small
\begin{tabular}{@{}lcc@{}}
\toprule
family set & transferred (pooled) & local (per-chapter) \\
\midrule
design (8)     & 0.0076\; (0.0\%) & 0.0776\; (37.5\%) \\
held-out (146) & 0.0383\; (6.9\%) & 0.1284\; (52.7\%) \\
\bottomrule
\end{tabular}
\caption{\textbf{Preregistered $2{\times}2$: selection vs.\ calibration-rate
transfer.} Method A \exact{} mean certified fraction (in-band share in
parentheses), min-eval subset, common $N{=}1000$.}
\label{tab:factorial}
\end{table}

\subsection{Does local calibration help?}
\label{sec:res-local}

\begin{figure}[t]
\centering
\includegraphics[width=0.62\textwidth]{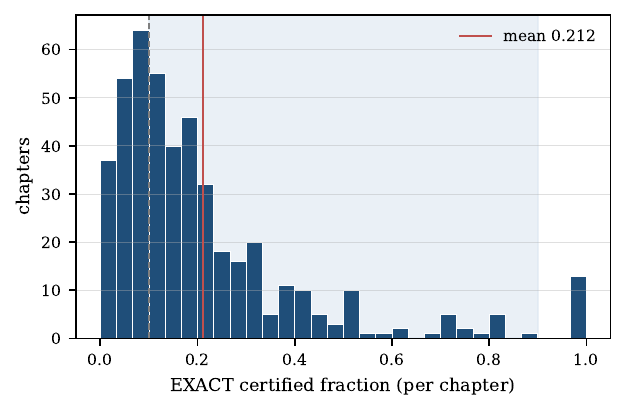}
\caption{\textbf{Scale distribution.} EXACT certified fraction across all 458
Missouri chapters (independent extractor, $N{=}1000$). Shaded: the
informative band; red line: the 0.212 mean.}
\label{fig:scale}
\end{figure}

Yes, materially. Under per-chapter calibration the held-out mean certified
fraction rises from 0.038 to 0.128 and the in-band share from 6.9\% to
52.7\% (Table~\ref{tab:factorial}, right column). The 304-chapter sweep under
the primary extractor tells the same story at distribution level: \exact{}
mean 0.134 with 53.6\% of chapters in band, and the full-corpus scale run
puts the \exact{} mean at 0.212 over all 458 chapters (0.168 on the
419-chapter min-eval subset, 63.3\% in band; Figure~\ref{fig:scale}). The
stability curve decays gracefully with noise scaling: 78.5\% of chapters in
band at half the measured rate, 49.9\% at twice it
(Appendix~\ref{app:stability}).

The 0.038 of Table~\ref{tab:heldout} and the 0.212 here are different
estimands, not a contradiction. Table~\ref{tab:heldout} measures
\emph{global deployment}: one error model, estimated on a disjoint
calibration half, transferred to chapters the method never saw, which is the
configuration a shipped tool would use by default.
Table~\ref{tab:factorial} varies exactly one factor of that design, the
calibration source, and Figure~\ref{fig:scale} measures the full corpus under
the primary extractor's in-corpus calibration. The gap between the two
numbers is precisely the price of rate transfer, which is why
Table~\ref{tab:factorial}, not either headline number alone, carries the
paper's second finding.

The operational consequence is the deployment doctrine: calibrate per chapter
or per family, or run tolerant with the $\varepsilon$ grid attached, and
report survival probability and edit distance next to every certified
implication. A single globally shipped exact error model is the one mode the
evidence rules out.

\subsection{Cross-jurisdiction replication}
\label{sec:res-india}

\begin{figure}[t]
\centering
\includegraphics[width=0.95\textwidth]{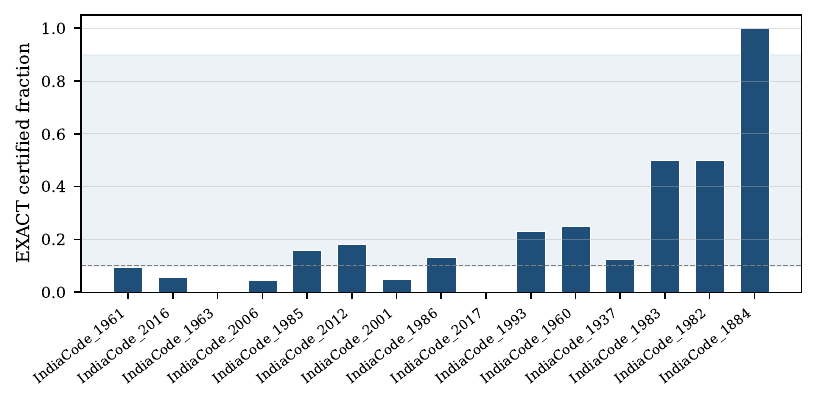}
\caption{\textbf{IndiaCode, per Act.} EXACT certified fraction (independent
extractor, $N{=}1000$). Tall bars are small-sample Acts; the
min-eval-eligible set lands in band 0/6.}
\label{fig:india}
\end{figure}

India fails, and we report it plainly. The pooled IndiaCode context (502
sections, basis 60, 57 evaluated) certifies \exact{} 0.000 under Method A
(\dstar{} 0.053). In the held-out-jurisdiction design, 6 of 8 Acts pass the
min-eval filter and \exact{} lands in band on \textbf{0 of 6} (mean 0.041).
Per-Act values span 0.0--1.0, but the top of the range is small-sample noise:
the 1.0 belongs to a 6-section Act with 3 evaluated implications
(Appendix~\ref{app:india}, Figure~\ref{fig:india}).

A plausible stressor is visible: cross-references are dense in IndiaCode
(\attr{xref} fires on 36.7\% of sections; 34.1\% resolve within the same
Act), and the Missouri ablations show the two cross-reference attributes are
load-bearing for certification (dropping either sends the pooled certificate
to 0.0; Appendix~\ref{app:ablations}). We did not isolate the cause in India
and do not claim one. What we claim is the fact: the certificate that passes
its Missouri gate does not transfer to this jurisdiction under the reported
setup.

\subsection{Controls: calibration, not noise volume}
\label{sec:res-controls}

Three negative controls establish that the certificate responds to measured
attribute-specific error rather than to generic noise
(Table~\ref{tab:controls}). Permuting the measured rates across attributes
(NC1) collapses and disjoints the certified set (Jaccard 0.0 with the
calibrated set on POOLED, ch375, ch115). An attribute-agnostic uniform flip
at the matched marginal rate (NC3) certifies 0.0 everywhere calibration
yields 0.017--0.067: which attribute errs matters, not how much error there
is. Marginal shuffling (NC2) shows the real bases (30/59/52 implications on
POOLED/ch375/ch115) are far smaller than marginal-matched random contexts
(151/133/125), so the target sets reflect joint statutory structure. The
independent DG cross-check agrees with NextClosure on every real context.

\begin{table}[t]
\centering\small
\begin{tabular}{@{}llccc@{}}
\toprule
control & quantity & POOLED & ch375 & ch115 \\
\midrule
NC1 & calibrated cert.\ frac.\ (baseline) & 0.067 & 0.017 & 0.020 \\
    & shuffled cert.\ frac.\ (mean) & 0.007 & 0.027 & --- \\
    & Jaccard(cert., shuffled) & 0.0 & 0.0 & 0.0 \\
\midrule
NC2 & real basis size & 30 & 59 & 52 \\
    & marginal-matched basis (mean) & 151.1 & 133 & 125 \\
\midrule
NC3 & matched global flip rate & 0.055 & 0.062 & --- \\
    & uniform-flip cert.\ frac. & 0.0 & 0.0 & 0.0 \\
\bottomrule
\end{tabular}
\caption{\textbf{Negative controls} (common $N{=}1000$). Dashes mark cells
the artifact leaves empty.}
\label{tab:controls}
\end{table}

\section{Discussion and Limitations}
\label{sec:discussion}

\subsection{Interpretation}

The instrument is usable but fragile, and both halves of that verdict carry
numbers.

Usable, because the certificate identifies a statistically supported core of
implications and makes failures auditable. It passes a preregistered held-out
gate it was allowed to fail, on exact and tolerant semantics, across 7 to 11
distinct Titles. The controls show the signal is measured calibration rather
than noise volume. The provenance layer turns every certified implication
into an auditable object and every fragile implication into a named, checkable
death: across the headline contexts, non-survival is overwhelmingly
instantiated violation rather than premise collapse
(Table~\ref{tab:decomposition}), so the killing row almost always exists
and can be read.

\begin{table}[t]
\centering\small
\begin{tabular}{@{}lccc@{}}
\toprule
context & $S_{\mathrm{nv}}$ & $S_{\mathrm{logic}}$ & collapse share \\
\midrule
POOLED & 0.517 & 0.517 & 0.000 \\
ch375  & 0.478 & 0.500 & 0.042 \\
ch301  & 0.524 & 0.524 & 0.000 \\
ch115  & 0.562 & 0.612 & 0.114 \\
ch67   & 0.502 & 0.505 & 0.005 \\
ch473  & 0.579 & 0.695 & \textbf{0.274} \\
ch407  & 0.509 & 0.518 & 0.019 \\
ch210  & 0.658 & 0.684 & 0.076 \\
ch143  & 0.697 & 0.701 & 0.013 \\
\bottomrule
\end{tabular}
\caption{\textbf{Survival decomposition} (Method A, measured rate).
$S_{\mathrm{nv}}$ counts only non-vacuous survivals; $S_{\mathrm{logic}}$
counts vacuous retention as survival; the gap is premise collapse (share of
non-survival).}
\label{tab:decomposition}
\end{table}

Fragile, on four measured axes. The calibration is inter-extractor
disagreement, so the certificate is conditional on that disagreement model.
A single global error model misrepresents heterogeneous chapters (93.2\%
below floor under global deployment; \S\ref{sec:res-factorial}).
Cross-jurisdiction transfer fails on IndiaCode (0/6;
\S\ref{sec:res-india}). And per-implication survival does not imply
basis-level stability: the mean whole-basis edit distance at the measured
rate is 21.0 across the 419-chapter scale run and peaks at 59.7 on ch407,
against a reference basis of 57 implications. Implication-level robustness
asks whether a given rule survives; basis-level stability asks whether the
basis as a document survives. The first is what the certificate delivers;
the second is what no deployment should assume.

The tolerant semantics needs the same care. The lift concentrates where
premise support is real (certified fraction 0.047 $\to$ 0.330 at support
20--49; 0.147 $\to$ 0.618 at support $\geq 50$; nothing below support 20,
because a 5\% allowance cannot absorb even one violation there), and at
$\varepsilon{=}0.20$ the pooled context certifies 0.967, above the
informativeness ceiling (Appendix~\ref{app:eps}). Enough tolerance
trivializes the instrument, so $\varepsilon$ stays a declared sensitivity
parameter, not a tuned forward path.

\subsection{Limitations}
\label{sec:limitations}

\begin{enumerate}[leftmargin=1.6em,itemsep=2pt]
\item \textbf{The reference extractor is not validated ground truth.} It is
documentary truth by construction inside the experiment. The certificate
bounds survival under parser disagreement; it does not bound the probability
that the extraction, or the implication, is legally correct.
\item \textbf{Disagreement may understate real error.} Both extractors are
rule-based and share the upstream operative-text split, the schema, and some
vocabulary; \attr{unless} and \attr{xref\_res} converge exactly. Converting
measured disagreement into validated extraction error needs a
construction-gold corpus or a third, differently built extractor, neither of
which we ran.
\item \textbf{Method A assumes per-attribute independence.} The \dstar{}
diagnostic, which preserves polarity-conditioned joint co-errors, does not
certify less (it matches or beats A in 8 of 9 headline contexts), but it is a
diagnostic, not a second independent extractor.
\item \textbf{Basis algorithm.} We use NextClosure rather than the originally
specified LinCbO/In-Close class; correctness is cross-checked against a
definition-based oracle and the \texttt{concepts} library on every real
context.
\item \textbf{India fails under the reported setup} (0/6 eligible Acts in
band). We do not generalize beyond Missouri, and the India failure is a
result, not an omission.
\item \textbf{The favourable core is modest.} The gate pass rests on 10
held-out families under global deployment; the instrument is weak in absolute
terms in that mode and should be deployed per-chapter-calibrated or tolerant.
\item \textbf{Protocol.} Preregistration is hash-frozen locally without a
third-party timestamped registry; the in-sample band analysis is
retrospective by construction, which is why the held-out design carries the
verdict.
\end{enumerate}

The correct reading of the certificate throughout is: a statistical lower
bound on implication survival under the calibrated inter-extractor
disagreement model. Anything stronger is outside the evidence.

\subsection{Deployment protocol}
\label{sec:deployment}

The evidence supports one concrete operating procedure and rules out one
other.

\begin{enumerate}[leftmargin=1.6em,itemsep=2pt]
\item \textbf{Calibrate per chapter or per family rather than globally.} Run both
extractors on the target chapter, estimate $p_{\mathrm{fp}}/p_{\mathrm{fn}}$
locally, and certify against that model. The factorial says this is the
difference between a 0.038 and a 0.128 held-out mean.
\item \textbf{If local calibration is impossible, run tolerant} and attach
the full $\varepsilon$ grid, reading $\varepsilon$ as a sensitivity axis. Do
not select $\varepsilon$ post hoc to reach a desired certified fraction.
\item \textbf{Ship the payload with each certified implication}: survival
probability, Wilson lower bound, minimal counterexample, likely
counterexample, and the context's whole-basis edit distance. A certified
implication without its counterexample is only half the instrument.
\item \textbf{Treat below-floor contexts as no-certificate zones.} Under a
transferred global model, most chapters are below the floor; the correct
output there is silence plus the diagnostic, not a weak certificate.
\item \textbf{Do not transfer across jurisdictions without recalibration.}
The IndiaCode result is the evidence that the certificate's validity is
jurisdiction-conditional.
\end{enumerate}

\section{Conclusion}
\label{sec:conclusion}

We set out to learn how much of the symbolic logic extracted from statutes
survives the extraction itself. The measured answer: a small, auditable core
survives with a certified lower bound on survival under the measured disagreement model, and the rest dies in checkable ways.
Support does not predict which is which; the attribute footprint of an
implication, crossed with the measured error structure of the extractor,
does. The survival certificate we built on that finding passes a
preregistered held-out gate on Missouri and fails cleanly on IndiaCode, and
the difference between its weak and strong deployment modes is not the
method but the calibration: per chapter, not global.

The deployment doctrine follows directly. Calibrate locally or run tolerant
with the grid attached; ship every certified implication with its survival
probability, edit distance, and minimal counterexample; treat the certificate
as a conditional reliability instrument. The open problems are the ones our
scope statements point at: validated extraction error via a
construction-gold corpus or a third extractor, and a cross-jurisdiction
account of why India resists the certificate Missouri admits.



\appendix
\section{Attribute schema}
\label{app:schema}

Table~\ref{tab:schema} defines the frozen 12-attribute schema exactly as the
reference parser implements it (\path{code/attributes.py}).

\begin{table}[H]
\centering\small
\setlength{\tabcolsep}{4pt}
\begin{tabular}{@{}lL{8.9cm}@{}}
\toprule
attribute & fires when the operative text contains \\
\midrule
\attr{oblig}     & \emph{shall}/\emph{must} imposing a duty, corrected so \emph{shall not} does not double-count \\
\attr{prohib}    & \emph{shall not}, \emph{may not}, \emph{must not}, \emph{no person shall} \\
\attr{permis}    & \emph{may} (excluding \emph{may not}) \\
\attr{proviso}   & a \emph{provided (however,) that} proviso marker \\
\attr{unless}    & the exception token \emph{unless} \\
\attr{except}    & \emph{except}, \emph{notwithstanding}, \emph{exception} \\
\attr{xref}      & a citation of another section or chapter (jurisdiction-specific pattern) \\
\attr{xref\_res} & an \attr{xref} whose cited id resolves within the same Act / corpus index \\
\attr{def}       & definitional language: \emph{means}, \emph{shall mean}, \emph{as used in}, \emph{includes} \\
\attr{numthresh} & a numeric quantity with a unit (digits or spelled numbers; \$, \%, days, dollars, inhabitants, \dots) \\
\attr{penalty}   & penalty language: \emph{fine}, \emph{imprison}, \emph{misdemeanor}, \emph{felony}, \emph{forfeit}, \dots \\
\attr{deadline}  & a numeric quantity with a time unit (\emph{within thirty days}, \dots) \\
\bottomrule
\end{tabular}
\caption{\textbf{The frozen 12-attribute schema.} All matching is over
operative text only; \attr{xref\_res} is act-scoped.}
\label{tab:schema}
\end{table}

\section{In-sample headline}
\label{app:headline}

Table~\ref{tab:headline} reports the nine development contexts. These
families were selected during development, so the numbers are descriptive;
the controlling verdict is the held-out design of
\S\ref{sec:res-heldout}.

\begin{table}[H]
\centering\small
\setlength{\tabcolsep}{3.5pt}
\begin{tabular}{@{}lrrrcccccc@{}}
\toprule
context & sec. & basis & eval & A \exact{} & A \exact{} FDR & A \textsc{tol}(.05) & \dstar{} \exact{} & \dstar{} \textsc{tol} & edit dist \\
\midrule
POOLED (8 Titles)        & 2555 & 30 & 30 & 0.067 & 0.033 & 0.633 & 0.033 & 0.700 & 34.9 \\
ch375 Insurance          &  383 & 59 & 59 & 0.017 & 0.017 & 0.034 & 0.017 & 0.068 & 57.0 \\
ch301 Motor Vehicles     &  297 & 38 & 38 & 0.158 & 0.158 & 0.421 & 0.237 & 0.579 & 29.7 \\
ch115 Elections          &  365 & 52 & 51 & 0.020 & 0.020 & 0.039 & 0.039 & 0.059 & 44.7 \\
ch67 Political Subdiv.   &  560 & 48 & 48 & 0.042 & 0.042 & 0.375 & 0.062 & 0.479 & 37.0 \\
ch473 Probate            &  251 & 47 & 42 & 0.167 & 0.167 & 0.190 & 0.238 & 0.262 & 29.6 \\
ch407 Merchandising      &  316 & 57 & 57 & 0.035 & 0.035 & 0.053 & 0.035 & 0.070 & 59.7 \\
ch210 Child/Pub.\ Health &  215 & 36 & 36 & 0.028 & 0.028 & 0.139 & 0.111 & 0.306 & 34.1 \\
ch143 Taxation           &  168 & 58 & 58 & 0.155 & 0.155 & 0.345 & 0.224 & 0.466 & 37.2 \\
\bottomrule
\end{tabular}
\caption{\textbf{Headline contexts, in-sample (exploratory, descriptive
only).} Certified fraction $=$ Wilson one-sided 95\% lower bound on survival
${\geq}\,\tau{=}0.95$, common $N{=}1000$, independent extractor, seed 0.}
\label{tab:headline}
\end{table}

\begin{figure}[H]
\centering
\includegraphics[width=0.95\textwidth]{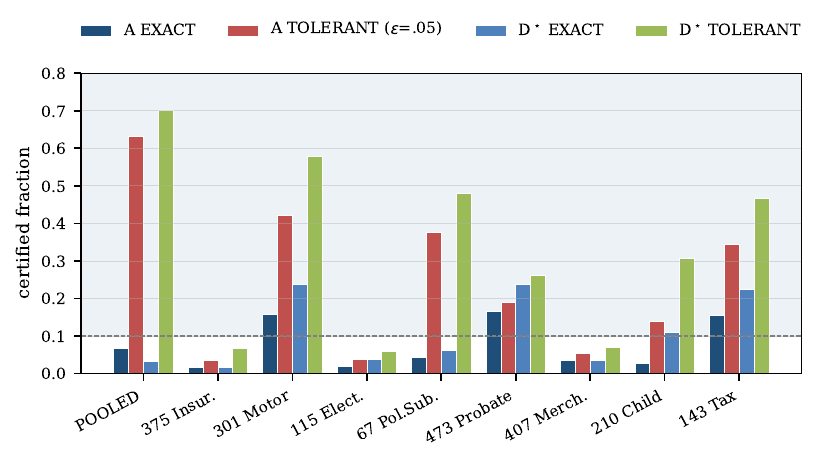}
\caption{\textbf{Headline contexts.} Certified fraction by estimator and
semantics. The shaded band is the preregistered informative region.}
\label{fig:headline}
\end{figure}

\section{Full $\varepsilon$ grid}
\label{app:eps}

\begin{table}[H]
\centering\small
\begin{tabular}{@{}lccccc@{}}
\toprule
context & $\varepsilon{=}0$ & .02 & .05 & .10 & .20 \\
\midrule
POOLED & 0.067 & 0.100 & 0.633 & 0.800 & 0.967 \\
ch375  & 0.017 & 0.017 & 0.034 & 0.271 & 0.627 \\
ch301  & 0.158 & 0.184 & 0.421 & 0.658 & 0.842 \\
ch115  & 0.020 & 0.020 & 0.039 & 0.196 & 0.333 \\
ch67   & 0.042 & 0.083 & 0.375 & 0.625 & 0.667 \\
ch473  & 0.167 & 0.167 & 0.190 & 0.286 & 0.357 \\
ch407  & 0.035 & 0.035 & 0.053 & 0.193 & 0.684 \\
ch210  & 0.028 & 0.028 & 0.139 & 0.361 & 0.556 \\
ch143  & 0.155 & 0.155 & 0.345 & 0.603 & 0.759 \\
\bottomrule
\end{tabular}
\caption{\textbf{Full $\varepsilon$ grid}, Method A certified fraction, all
headline contexts (common $N{=}1000$).}
\label{tab:eps}
\end{table}

\begin{figure}[H]
\centering
\includegraphics[width=0.62\textwidth]{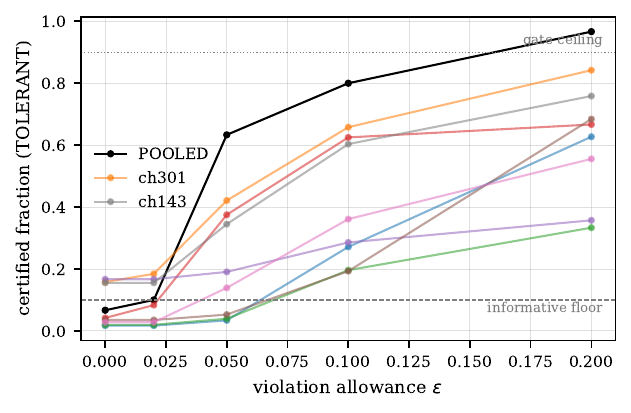}
\caption{\textbf{The $\varepsilon$ grid as curves.} Tolerance rescues dense,
high-support contexts first; at $\varepsilon{=}0.20$ POOLED exceeds the 0.90
informativeness ceiling.}
\label{fig:eps}
\end{figure}

\section{Per-attribute confusion matrix}
\label{app:confusion}

\begin{table}[H]
\centering\small
\begin{tabular}{@{}lcc@{}}
\toprule
attribute & $p_{\mathrm{fp}}$ & $p_{\mathrm{fn}}$ \\
\midrule
\attr{oblig}     & 0.1645 & 0.0053 \\
\attr{prohib}    & 0.1233 & 0.0000 \\
\attr{permis}    & 0.2109 & 0.0012 \\
\attr{proviso}   & 0.0030 & 0.0865 \\
\attr{unless}    & 0.0000 & 0.0000 \\
\attr{except}    & 0.0479 & 0.0000 \\
\attr{xref}      & 0.1373 & 0.0000 \\
\attr{xref\_res} & 0.0000 & 0.0000 \\
\attr{def}       & 0.1898 & 0.0215 \\
\attr{numthresh} & 0.0149 & 0.4294 \\
\attr{penalty}   & 0.0040 & 0.0030 \\
\attr{deadline}  & 0.0437 & 0.0000 \\
\bottomrule
\end{tabular}
\caption{\textbf{Measured per-attribute confusion} (POOLED, independent
extractor vs.\ reference). Inter-extractor disagreement, not validated
extraction error.}
\label{tab:confusion}
\end{table}

\section{Premise-support strata}
\label{app:strata}

\begin{table}[H]
\centering\small
\begin{tabular}{@{}lcccccc@{}}
\toprule
support & \#impl & $\bar{S}_{\mathrm{exact}}$ & $\bar{S}_{\mathrm{tol}}$ & cert.\ ex. & cert.\ tol. & lift \\
\midrule
1--9   & 148 & 0.634 & 0.634 & 0.074 & 0.074 & 0.000 \\
10--19 &  97 & 0.599 & 0.636 & 0.052 & 0.052 & 0.000 \\
20--49 & 106 & 0.522 & 0.778 & 0.047 & 0.330 & 0.283 \\
50+    &  68 & 0.392 & 0.777 & 0.147 & 0.618 & 0.471 \\
\bottomrule
\end{tabular}
\caption{\textbf{Premise-support strata}: mean survival and certified
fraction, \exact{} vs.\ \tolerant{} ($\varepsilon{=}0.05$), pooled over
headline contexts.}
\label{tab:strata}
\end{table}

\begin{figure}[H]
\centering
\includegraphics[width=0.62\textwidth]{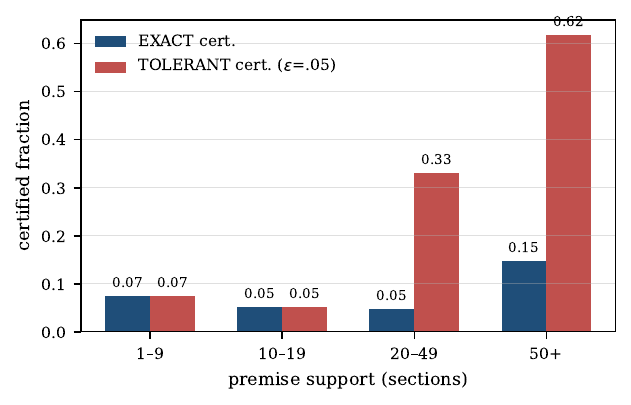}
\caption{\textbf{Premise support decides whether tolerance helps.} Below
support 20 the \tolerant{} and \exact{} certificates coincide; above it the
lift is large.}
\label{fig:support}
\end{figure}

\section{Survival decomposition}
\label{app:decomposition}

The survival decomposition is reported in the main text
(Table~\ref{tab:decomposition}) because it carries the interpretive claim
that implication deaths are checkable violations rather than support
artifacts.

\section{Stability under noise scaling}
\label{app:stability}

\begin{table}[H]
\centering\small
\begin{tabular}{@{}lccc@{}}
\toprule
$\lambda$ & mean cert.\ frac. & median & frac.\ in band \\
\midrule
0.0  & 1.000 & 1.000 & 0.000 \\
0.25 & 0.294 & 0.273 & 0.947 \\
0.5  & 0.213 & 0.184 & 0.785 \\
1.0  & 0.162 & 0.128 & 0.597 \\
2.0  & 0.144 & 0.098 & 0.499 \\
\bottomrule
\end{tabular}
\caption{\textbf{Aggregate stability curve}, all 458 chapters (independent
extractor): mean certified \exact{} fraction vs.\ noise scaling $\lambda$
($\lambda{=}1$ is the measured disagreement rate).}
\label{tab:lambda}
\end{table}

\begin{figure}[H]
\centering
\includegraphics[width=0.62\textwidth]{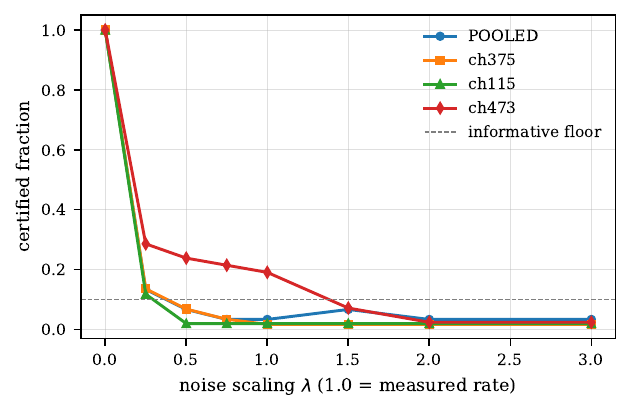}
\caption{\textbf{Stability under noise scaling}, four headline contexts. The
dashed line is the 0.10 informativeness floor.}
\label{fig:stability}
\end{figure}

\begin{figure}[H]
\centering
\includegraphics[width=0.62\textwidth]{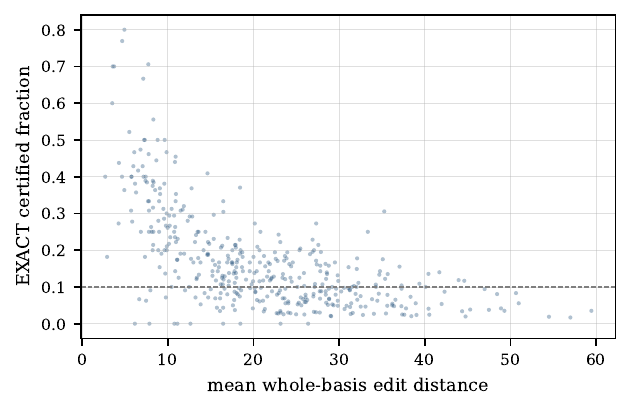}
\caption{\textbf{Basis-level movement vs.\ per-implication survival}, 419
min-eval chapters. High certified fraction does not imply a stable basis.}
\label{fig:editdist}
\end{figure}

\section{IndiaCode per-Act results}
\label{app:india}

\begin{table}[H]
\centering\small
\begin{tabular}{@{}lrrrrcc@{}}
\toprule
Act & sections & basis & $n_{\mathrm{eval}}$ & \attr{xref\_res} prev. & A \exact{} & A \textsc{tol}(.05) \\
\midrule
196125 & 84 & 38 & 32 & 0.452 & 0.094 & 0.094 \\
201618 & 69 & 41 & 36 & 0.304 & 0.056 & 0.056 \\
196345 & 64 & 28 & 23 & 0.234 & 0.000 & 0.000 \\
200635 & 59 & 47 & 43 & 0.305 & 0.046 & 0.046 \\
198513 & 41 & 29 & 25 & 0.463 & 0.160 & 0.160 \\
201213 & 38 & 23 & 22 & 0.368 & 0.182 & 0.182 \\
200145 & 38 & 30 & 21 & 0.289 & 0.048 & 0.048 \\
198602 & 36 & 29 & 23 & 0.333 & 0.130 & 0.130 \\
201722 & 18 & 24 & 19 & 0.333 & 0.000 & 0.000 \\
199333 & 13 & 19 & 13 & 0.462 & 0.231 & 0.231 \\
196064 & 11 & 15 &  8 & 0.273 & 0.250 & 0.250 \\
193701 & 11 & 19 & 16 & 0.273 & 0.125 & 0.125 \\
198313 &  7 & 13 &  8 & 0.143 & 0.500 & 0.500 \\
198201 &  7 & 13 &  8 & 0.143 & 0.500 & 0.500 \\
188412 &  6 & 11 &  3 & 0.500 & 1.000 & 1.000 \\
\midrule
INDIA\_POOLED & 502 & 60 & 57 & 0.341 & 0.000 & --- \\
\bottomrule
\end{tabular}
\caption{\textbf{IndiaCode replication}, acts by year--serial identifier.
Act-scoped \attr{xref\_res} throughout. Small Acts are high-variance; the
held-out design applies the $n_{\mathrm{eval}} \geq 10$ filter.}
\label{tab:india}
\end{table}

\section{Held-out in-band families}
\label{app:inband}

Table~\ref{tab:inband} lists the ten held-out Missouri chapters that carry the
controlling gate under \exact{} semantics, with their Titles. These are the
families on which the certificate stays informative even under the
global-deployment estimand; the remaining 136 eligible chapters fall below
the 0.10 floor.

\begin{table}[H]
\centering\small
\begin{tabular}{@{}lcrrccc@{}}
\toprule
chapter & Title & sections & basis & $n_{\mathrm{eval}}$ & A \exact{} & A \textsc{tol}(.05) \\
\midrule
130 & IX    & 42  & 41 & 40 & 0.1250 & 0.1500 \\
169 & XI    & 101 & 31 & 31 & 0.1290 & 0.1290 \\
170 & XI    & 45  & 33 & 29 & 0.1034 & 0.1034 \\
241 & XV    & 37  & 21 & 16 & 0.1250 & 0.1250 \\
261 & XVII  & 34  & 25 & 22 & 0.1364 & 0.1364 \\
276 & XVII  & 70  & 31 & 30 & 0.1000 & 0.1000 \\
281 & XVII  & 45  & 27 & 25 & 0.1200 & 0.1200 \\
523 & XXXVI & 34  & 33 & 30 & 0.1000 & 0.1000 \\
643 & XL    & 68  & 50 & 47 & 0.1064 & 0.1064 \\
700 & XLI   & 57  & 32 & 31 & 0.1290 & 0.1290 \\
\bottomrule
\end{tabular}
\caption{\textbf{The ten \exact{} in-band held-out chapters} (common
$N{=}1000$), spanning Titles IX, XI, XV, XVII, XXXVI, XL, XLI.}
\label{tab:inband}
\end{table}

\section{Ablations and sensitivity}
\label{app:ablations}

\textbf{Ablations} on POOLED: dropping \attr{xref} or \attr{xref\_res} sends
the certified fraction to 0.0; dropping \attr{numthresh} to 0.037; dropping
\attr{oblig} (basis shrinks 30 $\to$ 13) leaves 0.077. Fragility follows the
measured per-attribute error and the schema's load-bearing cross-references.

\label{app:sensitivity}
\textbf{Sensitivity.} The certification threshold moves the POOLED \exact{}
fraction through $\tau \in \{0.90, 0.95, 0.99\} \to \{0.133, 0.067, 0.033\}$;
the budget moves it through $N \in \{200, 500, 1000\} \to \{0.033, 0.033,
0.067\}$; the support floor $S_{\min} \in \{1,3,5,10\}$ leaves POOLED
unchanged. At $N{=}500$ the controlling gate also passes (4 \exact{} / 9
\tolerant{} in band).

\textbf{Seed variance.} Across seeds $\{0,1,2\}$ the certified-fraction
standard deviation is 0.000--0.0157 (max at POOLED: $\{0.067, 0.033,
0.033\}$). Same-seed reruns reproduce all survival counts exactly
(\texttt{run\_variance.py}: \texttt{all\_identical\_seed\_survival\_counts}).

\section{Audit trail and disowned estimators}
\label{app:audit}

This appendix records, as reproducibility documentation, the defects an
independent red-team review found in an earlier revision of this pipeline and
what the fixes changed.

\textbf{A1. Gate substitution.} The earlier revision evaluated a stricter
majority-in-band gate in place of the preregistered ${\geq}\,2$-family rule
and reported a ``confirmatory FAIL'' that was arithmetically inconsistent
with its own outputs (4 eligible \exact{} families in band against a
${\geq}\,2$ rule). The controlling gate is restored and passes (10 \exact{}
families / 7 Titles at $N{=}1000$; 4 \exact{} / 9 \tolerant{} at the
reviewer's $N{=}500$); the majority rule is reported as a post-hoc prevalence
diagnostic (fails on both semantics).

\textbf{A2. Confounded explanation.} The earlier analysis attributed the
held-out below-floor prevalence to post-selection generalization. The
preregistered $2{\times}2$ factorial (\S\ref{sec:res-factorial}) separates
the factors: rate transfer is the large effect, selection is small and
negative. The causal claim was replaced.

\textbf{A3. Invalid joint-error estimator.} The earlier Method D drew whole
donor error vectors and XOR-ed them unconditionally, manufacturing false
negatives on attributes whose measured $p_{\mathrm{fn}}$ is exactly 0
(\attr{prohib} 0.079, \attr{xref} 0.047, \attr{except} 0.032,
\attr{deadline} 0.030 injected). The resulting ``joint errors are more
destructive'' claim is \textbf{retracted}. The corrected
polarity-conditional \dstar{} applies donor residuals only where donor and
target reference polarity match; under it, \dstar{} ${\geq}$ A in 8 of 9
headline contexts (equal at ch375/ch407, lower only at POOLED;
per-implication correlation up to $\rho{=}0.97$), so joint co-error structure
is not more destructive than independent flips. The disowned estimator is
retained in the artifact as a labelled ablation, certifying ${\approx}\,0$
everywhere.

\textbf{Method C.} Section resampling of the fixed observed matrix; it
quantifies sampling uncertainty of the observed context and never gates
(headline range 0.263--0.600).

\textbf{Review provenance.} The red team reproduced the pipeline from a fresh
copy (0 SHA-256 mismatches, byte-identical headline outputs, zero non-timing
differences across scientific leaves) before issuing the findings above.

\end{document}